\documentclass{article}

\usepackage{lineno}
\usepackage{url}

\usepackage{booktabs}
\usepackage{framed}
\usepackage{scrextend}
\usepackage{bm}
\usepackage[mathscr]{eucal}
\usepackage{amsmath}
\usepackage{txfonts}
\usepackage{multirow}
\usepackage{changepage}
\usepackage{url}
\usepackage{graphicx}

\usepackage{color,soul}

\title{Mining Social Media for Newsgathering: A Review}

\author{Arkaitz Zubiaga \\ Queen Mary University of London, London, UK \\ a.zubiaga@qmul.ac.uk}

\date{}

\begin{document}

\maketitle

\begin{abstract}
 Social media is becoming an increasingly important data source for learning about breaking news and for following the latest developments of ongoing news. This is in part possible thanks to the existence of mobile devices, which allows anyone with access to the Internet to post updates from anywhere, leading in turn to a growing presence of citizen journalism. Consequently, social media has become a go-to resource for journalists during the process of newsgathering. Use of social media for newsgathering is however challenging, and suitable tools are needed in order to facilitate access to useful information for reporting. In this paper, we provide an overview of research in data mining and natural language processing for mining social media for newsgathering. We discuss five different areas that researchers have worked on to mitigate the challenges inherent to social media newsgathering: news discovery, curation of news, validation and verification of content, newsgathering dashboards, and other tasks. We outline the progress made so far in the field, summarise the current challenges as well as discuss future directions in the use of computational journalism to assist with social media newsgathering. This review is relevant to computer scientists researching news in social media as well as for interdisciplinary researchers interested in the intersection of computer science and journalism.
\end{abstract}

\section{Introduction}
\label{introduction}

The popularity of social media platforms has increased exponentially in recent years, gathering millions of updates on a daily basis \cite{fink2014data,fuchs2017social}. The use of ubiquitous digital devices to post on social media has in turn led to an increased presence of citizen journalism, where citizens, such as eyewitnesses on the ground, can share first-hand reports about an ongoing incident or event without having to be a professional who has access to journalistic media \cite{cohen2011computational,anderson2013towards}. The quintessential social media platform for citizen journalism, Twitter, has become a gold mine for journalists to learn about breaking news and to gather valuable information to broaden the coverage of their news reports \cite{knight2012journalism,knight2013social,belair2015social,abdenour2017digital}. Likewise, other major social media platforms such as Facebook and Instagram are increasingly used for newsgathering and reporting \cite{lysak2012facebook,murrell2014vulture,kaonza2015social,zakaria2018effect}, whereas platforms such as YouTube and Periscope give access to video content from eyewitnesses and practitioners on the ground \cite{ighiegba2018utilisation,fichet2016eyes,tang2016meerkat}. News agencies and journalists are increasingly turning to social media to gather the latest developments of news stories \cite{alejandro2010journalism,hermida2012social,broersma2013twitter,hermida2013journalism}. It is well known, however, that social media is a hodgepodge where users post all kinds of contents, ranging from irrelevant chatter to first-hand, eyewitness reports. While information such as the latter can be of great value, journalists need to sift through the vast amounts of updates posted in the stream of user-generated content to pick out the pieces of information of their interest, which tends to be costly, cumbersome, and often unfeasible for a human if done manually \cite{belair2015social}. To make the most of the stream of tweets for news reporting, journalists need bespoke tools that facilitate access to the relevant bits of information in a timely fashion. This has led to an increased interest in the scientific community to study techniques to facilitate newsgathering from social media.

\subsection{Social Media for Journalists}
\label{ssec:social-media-journalists}

Social media, while noisy, present numerous assets to the community of journalists and media professionals. Social media platforms such as Twitter, Facebook, Reddit, Instagram or YouTube, as tools that enable its users to share real-time updates about what they are observing or experiencing at the moment, have shown a great potential for discovering and tracking breaking news \cite{phuvipadawat2010breaking,hu2012breaking,gross2017harvesting,brands2018social,lewis2019social} as well as for more critical analyses of events that provide complementary context around events \cite{priya2019should}. Early examples that boosted Twitter's popularity as a tool for real-time news monitoring include the plane that landed in New York City's Hudson River back in 2009 \cite{lenhart2009twitter}, or the Arab Spring in 2011, which was largely discussed on Twitter \cite{khondker2011role}. This has in turn attracted the interest of journalists in the social media platform \cite{orellana2018attention,zubiaga2019social}, increasing their presence and becoming a ubiquitous tool to stay abreast of what is going on in the real, offline world.

With this increase in terms of popularity \cite{saldana2017sharing}, social media has become a powerful tool for journalists and news organisations ranging from health journalism \cite{shoenberger2017perceived} to sports journalism \cite{li2017better}, including emergency journalism \cite{bowdon2014tweeting} and political journalism \cite{parmelee2013political}. Social media in the context of journalism is being mainly used in the following three ways: (i) as a venue to freely post news stories to reach out to potential new readers and increase the number of visitors on their websites, cf. \cite{kwak2010twitter}, (ii) as an analytical platform to explore the preferences of news consumers, analysing the news stories that users read and share most, cf. \cite{diakopoulos2014newsworthiness}, and (iii) as a gold mine to catch the scoop on breaking news, to retrieve additional context to broaden the coverage of their news reports (Muthukumaraswamy, 2010), and to reach out potential eyewitnesses who they might want to interview \cite{diakopoulos2012finding}. Here we focus on the latter, studying the different directions in which social media mining techniques are being developed and researched to facilitate newsgathering from social media.

When social media is used for newsgathering, however, it presents the challenge that the stream of updates flows much faster than a human can follow, with hundreds or even thousands of posts per minute, which makes it impossible for a human to keep track of everything that is being said. There is indeed the need to curate all these contents in such a way that the end user can follow \cite{imran2013extracting,zubiaga2013curating,tolmie2017supporting}. Journalists are in great need of applications that facilitate newsgathering work \cite{thurman2018social}. The user, be it a journalist or another news practitioner, can be interested in certain types of information, such as eyewitness reports, new reports that add to a developing story, contradicting reports that clarify previous claims, etc., while getting rid of redundant information.

\subsection{Scope and Structure of the Survey}
\label{ssec:structure}

\begin{figure}[tbh]
\centering
\includegraphics[width=\textwidth]{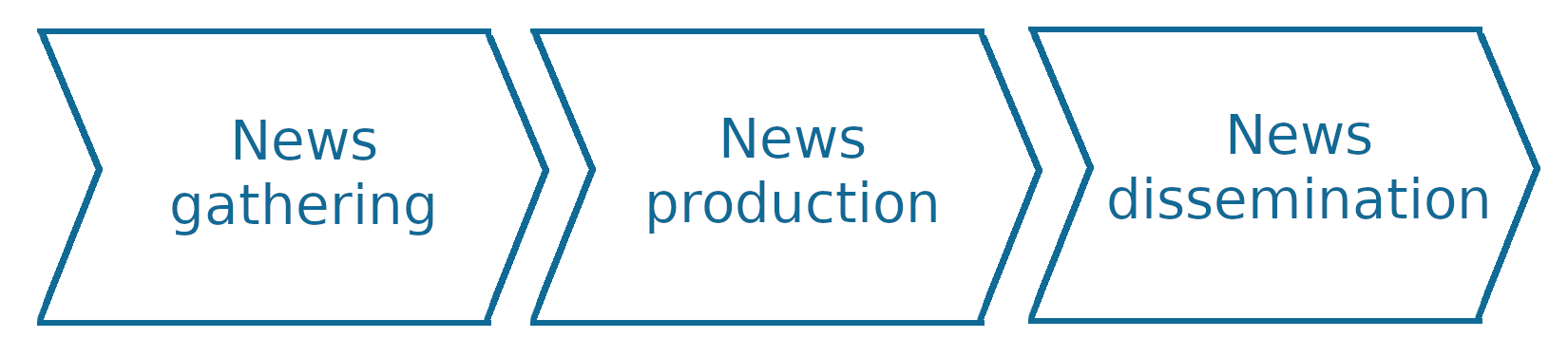}
  \caption{News cycle in journalism, including newsgathering (which is covered in this survey), news production and news dissemination.}
  \label{fig:news-cycle}
\end{figure}

The news cycle in journalism is a process consisting of three key stages \cite{hermida2010tv} (see Figure \ref{fig:news-cycle}): (1) gathering, where news stories are identified and researched for contextualisation, (2) production, where the information gathered in the preceding stage is collated into a news story for the medium in question, and (3) news dissemination, where the news story is distributed through media including TV, radio, online and paper. This review paper concerns the first stage of the news cycle, i.e. newsgathering, particularly focusing on social media as a source that has been increasingly studied in recent years for discovery and research around news.

The survey is organised in five different sections relating to different tasks within the use of social media for newsgathering:

\begin{itemize}
 \item \textbf{news discovery}, including event detection and news recommender systems,
 \item \textbf{curation of news}, including summarisation and finding information sources such as eyewitnesses or experts,
 \item \textbf{validation of content}, focusing on verification of content,
 \item \textbf{newsgathering dashboards}, discussing development of prototypes and tools to facilitate newsgathering, and
 \item \textbf{other tasks}, where more specific applications are discussed.
\end{itemize}

Further, we set forth future directions of research in the field by looking at current gaps.

The survey is intended to cover research regarding the use of computational methods for the purposes of newsgathering in journalism. It is not intended as an exhaustive survey in each of the NLP tasks discussed from a broader perspective, e.g. discussing all existing approaches to event detection, but rather delving into those that have shown a clear focus on journalistic purposes. Where there are surveys covering each specific task from a broader perspective and not focusing on journalistic purposes, these are referenced as recommendations for further reading.

\section{News Discovery}
\label{sec:news-discovery}

\subsection{Newsworthy Event Detection}
\label{ssec:event-detection}

An important task for supporting news discovery from social media is event detection, i.e. tracking the stream of updates from social media to identify newly emerging, newsworthy events. The objective of an event detection system in the context of journalism is to generate early alerts of candidate stories to be reported. Dou et al. \cite{dou2012event} provide an overview of the event detection task. They use the term ``new event detection'' instead for the case in which events that are not known a priori are to be identified, as opposed to detection of scheduled or predictable events. They define the new event detection task as that consisting in ``identifying the first story on topics of interest through constantly monitoring news streams.'' Further, they distinguish two different ways of tackling the new event detection (NED) task: (1) retrospective NED, where incoming social media posts are compared with past data to determine if new data is likely to be associated with a new event, and (2) online NED, where incoming social media posts are clustered online, where each new post is either included in an existing cluster (event) or, when it is different to existing clusters, it makes its own cluster, i.e. a new event. Once an event is identified, they propose to develop additional components to track and summarise those events.

While most work on event detection uses social media as the main data source, some have proposed to leverage other data sources to then propagate to social media. This is the case of Twitcident, proposed by Abel et al. \cite{abel2012twitcident}, which uses emergency broadcasting services to identify incidents, extract a set of related keywords, and then collect associated updates from social media. This approach is however dependent on those broadcasting services, and does not allow for early detection of events where broadcasting services or government officials may delay. Indeed, for this reason, most of the recent work in real-time event detection has instead used social media as the main data source. An early approach to breaking news detection from social media was introduced by Phuvipadawat and Murata \cite{phuvipadawat2010breaking}, who proposed to group tweets by similarity; in a subsequent step, these groups could be ranked by their probability to be linked to a newsworthy event. Given that a simple grouping approach based on similarity as overlap of all tokens in tweets would lead to noisy groups, they proposed instead to identify named entities, which are used to compute similarity scores and create clusters from a stream of tweets. They built a dataset combining tweets from a generic stream as well as tweets including the hashtag ``\#breakingnews''. This is a clever, simple approach to the problem, which proved to be effective in identifying news reports. However, later work has focused on identifying small scale incidents and news, which is more challenging owing to the small volume of data associated with the events.

Work on small-scale or local incidents and news stories attracted a substantial interest in the following years. Schulz et al. \cite{schulz2013see} developed an approach consisting of two steps: (1) automatic classification of user-generated content related to small scale incidents, and (2) filtering of irrelevant content, using Stanford NER for named entity recognition which enabled them to identify spatial, temporal, and thematic aspects. Weiler et al. \cite{weiler2013event} followed a similar approach to analyse location and temporal aspects of social media posts. They proposed to use a metric calculated by combining the log-likelihood ratio for geographic and temporal dimensions, which they used for identifying events in the content of tweets located in predefined local areas. Use of location-based features has indeed proven useful for breaking news detection through social media. A study by Wei et al. \cite{wei2017finding} focused on this aspect. They proposed an approach to estimate the location of Twitter users based on the locations of their friends. Using social media data enriched with this location information, they showed significant improvements in detection of local news as opposed to a baseline system that used Twitter's existing geolocation features.

A state-of-the-art approach to breaking news detection from social media is TopicSketch, introduced by Xie et al. \cite{xie2016topicsketch}. TopicSketch is intended to detect bursty topics in a real-time fashion from a large-scale data source like Twitter. It performs two steps: (1) a topic modelling step, where it looks for the acceleration of words and pairs of words over time, and (2) a hashing-based dimensionality reduction step, where only tweets observed in the stream in the last 15 minutes are stored; as this is still too large, it hashes all those words into buckets, to then identify frequent buckets.

Another system for real-time detection of news is Reuters Tracer \cite{liu2016reuters}. The system cleans up a stream of tweets through a number of filters: language filter, tweet-level spam filter, entity extraction and chit-chat filter. The cleaned stream is then fed into a clustering system that serves as a proxy for detecting stories, classifying them by topic and ranking them by newsworthiness.

While event detection applied to the specific context of news and journalism has been rather limited, there has been a substantial body of work looking at detection of other kinds of events, not necessarily newsworthy or of interest to news practitioners. For instance, a body of work has looked at event detection from social media for disaster management and to improve situational awareness \cite{sakaki2010earthquake,rogstadius2013crisistracker,middleton2014real,avvenuti2014ears,yin2015using,maresh2016social,avvenuti2016impromptu,imran2018processing,toujani2019event} or more generally at detecting trending topics in social media \cite{asur2011trends,becker2011beyond,aiello2013sensing,pervin2013fast,zubiaga2015real,naveed2019location}. Likewise, there are a number of applications that have been developed for disaster mapping, despite not always being documented in scientific papers, such as Ushahidi\footnote{\url{https://www.ushahidi.com/}} and Crowdmap\footnote{\url{https://crowdmap.com/}}. For a broader perspective on event detection methods for social media, beyond their application to journalism and news, see the detailed review by \cite{atefeh2015survey}.

\subsection{News Recommenders}
\label{sec:news-recommenders}

News recommenders are intended for journalists to identify top stories discussed in social media. This enables journalists to choose a story to cover based on people's interests. These recommendations can however be personalised, especially for journalists who cover specific subjects or types of journalism. Work in this direction is scarce. The first, preliminary approach to recommending news stories from social media in real-time was described by Phelan et al. \cite{phelan2009using}. They compared three different ranking algorithms, which showed promising results through preliminary experimentation. The same authors documented more advanced work in \cite{phelan2011terms}. They combined two data sources, tweets and RSS feeds from news outlets, to look for overlapping content. The system then ranks Twitter content overlapping with RSS feeds based on the friends of the user in question, recommending the top-ranked news items from Twitter. The resulting ranking of news is personalised for the user of the system, as it incorporates information from their network.

With a slightly different objective, Krestel et al. \cite{krestel2015tweet} described a system that, given a news story as input, recommends relevant tweets for gathering furthering evidence about the story from Twitter. They use language models and topic models to identify tweets that are similar to the news story given as input.

Another recommended system intended for newsgathering is called ``What to Write and Why (W$^3$)'', introduced by Cucchiarelli et al. \cite{cucchiarelli2017write,cucchiarelli2019topic}. The intuition of W$^3$ is to find stories that are being discussed by users in social media, which are yet to be covered by news outlets. To achieve that, they combine three different data sources: a stream of tweets, a stream of Wikipedia page views and a stream of news articles. A trending topic on Twitter and Wikipedia which is not present in news articles reflects a story of potential interest to be covered. They follow four steps: (1) event detection, where the SAX* algorithm is used for detecting events in each data source, (2) intra-source clustering, where events detected in each data source are clustered, (3) inter-source alignment, where clusters across different data sources are aligned, and (4) recommendation, where clusters are ranked to then recommend the top stories to the user. A similar approach to recommending uncovered news is presented by \cite{stilo2017capturing}.

\section{Curation of News}
\label{sec:news-curation}

\subsection{Summarisation}
\label{ssec:summarisation}

Another task that can assist with social media newsgathering is summarisation, i.e. automatically producing summaries for streams or timelines associated with a particular news story. Summarisation from social media is generally considered as a multi-document summarisation task, i.e. where multiple social media posts are given as input, and the system needs to produce a concise summary by getting rid of redundant information across posts. There are two main factors that vary across existing social media news summarisation approaches: (1) the type of summary can be extractive, where the summary is made of excerpts of the original text, or abstractive, where the summary is a new, rewritten text, and (2) summarisers have been proposed for scheduled events, such as sporting events where the vocabulary that will be used and the expected volume of posts can be predicted, and for unexpected events, such as breaking news where there is no prior information and the summariser needs to deal with new, unseen events.

Most existing work has focused on extractive summarisation. For scheduled events such as football matches, both \cite{zubiaga2012towards} and \cite{nichols2012summarizing} showed that it is relatively easy to achieve highly accurate summaries including all major happenings, such as goals and red cards, by simply looking at word frequencies. Indeed the vocabulary is largely predictable for football matches, and the volume of posts associated with keywords such as `goal' increases drastically when a team scores. \cite{khan2013multi} proposed an alternative, graph-based approach. First, they used LDA to identify topics of discussion within a stream. Then, they built a graph with the tweets in each topic, based on word co-occurrences. Finally, they used PageRank to identify salient tweets in each topic. Using a dataset of tweets associated with a presidential debate, they showed that their approach outperformed a word frequency based approach that used TF-IDF for weighting. \cite{meladianos2018optimization} proposed another graph-based approach, in this case analysing rapid changes in the graphs as a proxy to identify important events to be included in the summary. Using a graph representation of incoming tweets, their system uses a convex optimisation method to detect rapid changes in the weights of the edges. Representative tweets for those detected events are then extracted for the summary. Again, their experiments focused on sporting events.

Another approach introduced by \cite{wang2015summarization} focused on efficient and effective clustering of incoming tweets. They proposed Sumblr, a streaming clustering approach that assigns a cluster to incoming tweets. Incrementally, candidate tweets that will be part of the summary are selected for each of the clusters. As the number of clusters can grow indefinitely, the algorithm stores the timestamp of the last update of each cluster, which is used to delete outdated clusters that have not been updated for a while. The algorithm, intended to run in real-time while tweets are coming, showed to be highly efficient compared to other stream clustering techniques.

Work on both abstractive summarisation systems and for unexpected events such as breaking news is still in its infancy. To date, the only documented system with these characteristics is that by \cite{shapira2017interactive}. They first perform syntactic parsing of the tweets to create graphs connecting the predicates and arguments in the sentences. These graphs are further enhanced mention links, argument links and entailment links. They then produce scores for each mention in the texts, which enable them to select the mentions (entities and predicates) that will form the final, abstractive summary. While the quality of the summaries was not evaluated quantitatively, they did usability tests where participants rated different aspects of the summarisation system, showing promising results.

\subsection{Finding Information Sources}
\label{sec:information-sources}

It is not only the content that is important in the newsgathering process. Journalists need to identify quality information sources they can follow during an event \cite{starbird2012learning,morstatter2014finding,diakopoulos2012finding}. This information sources can be eyewitnesses on the ground, local people who know the area where the news is developing, experts in the topic, etc. They may want to follow them to get the latest updates, but they may also want to contact them for an interview or for further information.

\cite{diakopoulos2012finding} introduced Seriously Rapid Source Review (SRSR), a system that classifies social media users into one of the following categories: organisations, journalists/bloggers and ordinary people. They used a KNN classifier with five types of features to classify users during the 2011 England riots: network features, activity features, interaction features, named entities and topic distribution. The classifier achieved high precision (0.89) whereas the recall was relatively low (0.32). The authors argue that this is acceptable as journalists want to identify some users with high certainty (precision) while it is not necessary to identify them all (recall). \cite{starbird2012learning} developed a similar classifier, in this a binary classifier distinguishing users on the ground and not on the ground. A difficult challenge they had to deal with was the fact that less than 5\% of the users are actually on the ground, while the rest are tweeting from elsewhere, which makes the classification challenging. They implemented an SVM classifier that used several activity features, such as follower growth throughout the event as well as the number of retweets they got. With experiments on a dataset collected during the Occupy Wall Street protests, their classifier was able to detect, on average, 68\% of users on the ground. \cite{mahata2012does} had a slightly different objective of analysing event-specific sources that provide useful information for reporting purposes. They proposed two different features for their classifier: specificity and closeness.

Others have formulated to source-finding task as a geolocation problem. In this case, the objective is to identify whether users are located within the region affected by the crisis event in question. A binary classifier would then determine if a user is within a distance of the event or not. \cite{morstatter2014finding} described a classifier that determined if a user was IR (inside the region) or OR (outside the region). They proposed to look only at the linguistic features of the content posted by the users. By using linguistic features like unigrams, bigrams and part-of-speech tags, and a Naive Bayes classifier, they achieved F1 scores of 0.831 for the 2013 Boston Bombing and 0.882 for the 2012 Hurricane Sandy.

Others have specifically addressed the task of identifying witnesses, i.e. users who report to be on the ground and sharing first hand information of the event. This is indeed a very important task, and of great interest to journalists and newsrooms. However, the main challenge lies in the very small number of users who can be classified as witnesses. Indeed, for major events discussed worldwide, the ratio of witnesses tends to be very low. Exploring a range of crisis events, \cite{fang2016witness} found 401 witness tweets and 118,700 tweets, which leads to a ratio of 0.3\% witness tweets. For the classification of witness vs non-witness tweets, a range of features have been used, especially focusing on linguistic features, such as use of first-personal pronouns, temporal markers, locative markers or exclamative punctuation \cite{doggett2016identifying}. Both \cite{fang2016witness} and \cite{doggett2016identifying} reported high performance scores of above 80\% in terms of F1 score. While their experiments were preliminary, more work is still needed to achieve generalisation of these results to new, unseen events.

\cite{truelove2014testing} provided a taxonomy of types of accounts on what they called the ``Witness Model''. They define four types of accounts: (1) witness accounts, or accounts reporting a direct observation of the event or its effects, (2) impact accounts, or those being directly impacted or taking direct action because of the event and/or its effects, (3) relay accounts, in which a micro-blogger relays a witness or impact account, and (4) other accounts, which are users relevant to the event, but do not fall in any of the other three categories. Besides textual content, \cite{truelove2017identifying,truelove2017testing} used geotags and image features to perform supervised classification of users as witnesses or non-witnesses. They used unigrams, bigrams and part-of-speech tags as textual features. They showed that the image-based features performed substantially better (precision = 0.978, recall = 0.943) than textual features (precision = 0.803, recall = 0.588) in detecting witness accounts. In line with previous research, they also emphasise the difficulty of the task provided the small ratio of witness accounts with respect to non-witness accounts. One limitation of the work by \cite{truelove2017identifying} is that they only analyse geolocated tweets, which is a small fraction of the entire Twitter stream, and could therefore be missing out important updates from non-geolocated, witness users.

Given the small ratio of witness users reported in the work above, research into finding eyewitnesses from unfiltered streams has not yet been conducted. A good approximation is that by \cite{starbird2012learning}, who looked into social signals as potentially indicative of a user being a witness. Instead, others have looked into the content of posts. Research comparing these two approaches would also help determine future directions in the detection of eyewitnesses.



\subsection{Validation and Verification of Content}
\label{sec:content-verification}

As well as being a useful source of unique information in the early stages of emergencies and breaking news reports, one problem with social media is that it is easy for anyone to post fake reports, which leads to the streams of information associated with breaking news being rife with inaccurate information \cite{silverman2014verification,gupta2013faking}. It is of paramount importance for journalists to make sure that the reports they collect from social media are accurate \cite{schifferes2014identifying,brandtzaeg2016emerging,fletcher2017building}, given the bad consequences that reporting inaccurate news may have \cite{lewandowsky2012misinformation}, not only for the reputation of the news outlet but also for citizens. Aiming to analyse the impact of the spread of misinformation in social media, researchers have looked into the ability of ordinary social media users to discern misinformation from accurate reports. Using a crowdsourcing platform to run user studies where users were asked to determine if a piece of information was correct, \cite{zubiaga2014tweet} showed that the tendency is for users to assume that most of the information is accurate. This raises concern about the inability of ordinary users to detect misinformation. Research in recent years has focused on developing automated verification systems.

\cite{liu2015real} extended previous work by \cite{yang2012automatic} and \cite{castillo2011information} by incorporating what they called ``verification features''. These features were determined based on insights from journalists and included source credibility, source identification, source diversity, source and witness location, event propagation and belief identification. The authors showed that the proposed approach outperformed the two baselines. \cite{vosoughi2015automatic} tackled the verification task using three types of features: linguistic, user oriented and temporal propagation features. They compared two classifiers: Dynamic Time Wrapping (DTW) and Hidden Markov Models (HMMs). The results showed that HMMs outperformed DTWs. Temporal features proved to be the most helpful features in their experiments. \cite{boididou2017verifying} performed a comparison of three verification techniques: (1) use of textual patterns to extract claims about whether a tweet is fake or real and attribution statements about the source of the content; (2) exploiting information that same-topic tweets should also be similar in terms of credibility; and (3) use of a semi-supervised learning scheme that leverages the decisions of two independent credibility classifiers. Experiments on a wide range of datasets, they found the third approach performed best.

Verification has also been tackled as a task associated with rumours that spread in social media. This is the case of the RumourEval shared task at SemEval 2017 \cite{derczynski2017semeval}. Subtask B consisted in determining if each of the rumours in the dataset were true, false or remained unverified. The best approach, by Enayet and El-Beltagy \cite{enayet2017niletmrg}, used a stance classification system to determine the stance of each post associated with a rumour, which are aggregated to determine the likely veracity of the rumour. The verification task has sometimes been also framed as a fact-checking task, in which fact-checkable claims are checked against a database for accuracy. Most work on fact-checking claims has built knowledge graphs out of knowledge bases, such as Wikidata, to check the validity of claims \cite{wu2014toward,ciampaglia2015computational,shiralkar2017computational,shi2016discriminative}. Use of fact-checking techniques can be of limited help in the context of breaking news events, where much of the information is new and may not be available in knowledge bases. Others, such as Shao et al. \cite{shao2016hoaxy} with Hoaxy, implemented a dashboard for tracking the spread of misinformation, which could be used to better understanding this phenomenon and to get new insights for improving verification systems.

While this review mainly focuses on verification of textual content associated with news, there has been work on multimedia verification, including images and video. For more details on this line of research, the reader is referred to \cite{boididou2015verifying,zampoglou2016web,middleton2016geoparsing,middleton2018social} (image) and \cite{andreadou2015media,papadopoulou2017web} (video).

Research in verification has increased substantially in recent years with the advent of ``fake news'' as a major phenomenon in society, and given that the focus of this review article is much broader and focuses widely in newsgathering from social media, a comprehensive review of verification approaches is not within our scope. For further discussion on verification it is recommended to read detailed survey papers on rumour detection and resolution \cite{zubiaga2018detection} and fake news detection \cite{shu2017fake}, both in the context of social media.


\section{Newsgathering Dashboards}
\label{sec:dashboards}

Newsgathering from social media is also a topic of interest in the HCI research community. There have been numerous attempts at putting together dashboards specifically designed for newsgathering from social media. The first such application was TwitInfo \cite{marcus2011twitinfo}. TwitInfo was intended to allow journalists to track the latest developments associated with events in a real-time fashion, as opposed to post-hoc analysis, which is crucial for journalists looking for early reports. They treat the tweet stream as a signal processing problem, binning the frequencies (e.g. per minute) to identify peaks that are indicative of subevents. As frequencies evolve over time depending on the popularity of a event, TwitInfo continually updates mean frequency values observed so far. This in turn modifies the threshold for a new bin to be considered a peak and hence a subevent. While this step is designed as an event detection system, they then implemented a dashboard that visualised, in real-time, the subevents associated with an event. The dashboard was tested with 12 participants, proving it a useful application for data analysis for journalists. Similar approaches have been followed by others. Diakopoulos et al. \cite{diakopoulos2012finding} developed a dashboard for exploration of users to identify different types of users reporting about an event. Zubiaga et al. \cite{zubiaga2013curating} developed a dashboard for detailed exploration of news reports extracted from Twitter's trending topics. Lin et al. \cite{lin2016information} developed a similar application, whose visualisations largely focus on the social network and interactions between users during breaking news events, rather than the content of the tweets, which is the focus of \cite{zubiaga2013curating}. All these works showed the potential of Twitter as a newsgathering platform with the assistance of bespoke tools. Another dashboard, called MARSA, was introduced by Amer-Yahia et al. \cite{amer2012maqsa}. Their dashboard also enables visualisation of the temporal evolution of events, showing the most important news articles and tweets, mapped information, and the main entities being mentioned.

Schwartz et al. \cite{schwartz2015editorial} focused instead on geolocated social media posts when they developed CityBeat, a dashboard for exploration of local news reports. Having deployed their application in several newsrooms, an important limitation they found is the popularity bias. Algorithms gathering news from social media tend to favour popular news stories that were discussed by a significant number of people, missing out on the less popular, long tail of news stories of interest to fewer people.

Tolmie et al. \cite{tolmie2017supporting} performed a comprehensive study into the use of dashboards for newsgathering from social media, particularly focusing on the problem of verifying content by using dashboards. Through user studies in a newsroom, the authors concluded with reflections on the design of tools to support journalistic work with social media. Among others, the authors highlight the importance of time pressure, temporal cycles, and deadlines, hence suggesting that algorithms for newsgathering need to be both accurate and efficient. Other important findings included considering aspects specific to newsrooms, as the audiences, national and legal frameworks, and the type of editorial control, vary across newsrooms, and require journalistic tools to take those into consideration.

\section{Other Tasks}
\label{sec:other-tasks}

There have been numerous other attempts at mitigating the challenges of newsgathering from social media. We list and briefly discuss these other tasks in this section.

\textit{\textbf{Event monitoring.}} Event tracking through social media is usually performed by predefining a set of keywords or hashtags to collect the data. The list of keywords may however vary over time, and may need to be expanded in order to ensure collection of relevant data. There has been research looking into keyword expansion for these purposes. Improved collection of event-related tweets for increased coverage was pioneered by Milajevs et al. \cite{milajevs2013real}. They used Latent Dirichlet Allocation (LDA) to enable discovery of relevant keywords associated with a known event, which was designed for and tested with leisure events, which in their case included music and sporting events. Its application to other kinds of emergent, unpredictable events was not studied. Wang et al. \cite{wang2013exploiting,wang2014identifying,wang2015adaptive} proposed an adaptive microblog crawling approach to detect new hashtags linked to the event of interest. Their approach looks at correlations between hashtag co-occurrences to identify those that exceed a threshold. The limitation of this approach is that it is only applicable to hashtags, and largely dependent on a manually predefined threshold. Fresno et al. \cite{fresno2015exploiting} proposed to use features derived from user metadata, geolocation and time to overcome the limited availability of textual content. Their approach led to improved results, however showing difficulty to generalise to new events.

\textit{\textbf{Extracting information nuggets during crises.}} Imran et al. \cite{imran2013extracting} define information nuggets as brief, self-contained information items relevant to disaster response. They first classify tweets as personal, informative or other. Then, they define a typology of five types of informative tweets: (1) caution and advice, (2) casualties and damage, (3) donations of money, goods or services, (4) people missing, found or seen, and (5) information source. They implemented a Naive Bayes classifier with a set of linguistic and social features, achieving a precision of 0.57 and a recall of 0.57.

\textit{\textbf{Identifying news from tweets.}} Freitas et al. \cite{freitas2016identifying} aimed to identify newsworthy tweets, i.e. tweets that are worthy of news reporting. Using two Twitter datasets, they introduced an active training approach to rank tweets by newsworthiness, achieving an F1 score of 0.778.

\textit{\textbf{Classification of tweets.}} There has been work classifying disaster-related tweets. Stowe et al. \cite{stowe2016identifying} introduced an annotation schema with seven types of tweets: (1) sentiment, tweets expressing emotions towards the event, (2) action, describing actions for event preparedness, (3) preparation, describing preparation plans, (4) reporting, which includes first-hand information, (5) information, sharing or seeking information from others, (6) movement, which refer to sheltering or evacuation, and (7) others. They used an SVM classifiers with a range of features including word embeddings, part-of-speech tags and social features, achieving an F1 score of 0.551.

\textit{\textbf{Stance detection.}} Recent work has proposed to look into the stance expressed by different users in social media towards a particular piece of information that is circulating. The intuition behind this is that the aggregated stance expressed by users could be indicative of the veracity of the information, which could be leveraged for verification purposes. Lukasik et al. \cite{lukasik2016hawkes} proposed to use a Hawkes Process classifier to leverage the temporal evolution of stances. Research has shown that leveraging the temporal component and the evolving discourse around news stories is important to determine the stance of individual posts. Zubiaga et al. \cite{zubiaga2018discourse} conducted a comprehensive study comparing different sequential classifier to mine the discursive structure of social media information, finding that a Long-Short Term Memory network (LSTM) performed best, outperforming Conditional Random Fields (CRF) and Hawkes Processes.

\textit{\textbf{First story detection.}} The aim of the first story detection task is to find the earliest social media post (e.g. tweet) that reported a particular story. Work in this direction has explored the use of locality-sensitive hashing to be able to perform at scale \cite{petrovic2010streaming}, using paraphrases to detect sentences phrased differently \cite{petrovic2012using} and used moderated sources such as Wikipedia to avoid false positives in story detection from social media \cite{osborne2012bieber}.

\textit{\textbf{Identifying high-consensus news.}} Babaei et al. \cite{babaei2018purple} build on the idea that, for many news stories, there are two groups of users with different opinions, which they label red and blue. Looking at stances from social media users, who may agree or disagree with a news article's content, they aim to identify news stories where there is a consensus across different user groups.

\section{Discussion and Future Research Directions}
\label{sec:discussion}

In this review we have outlined the efforts in the scientific community to tackle the increasingly important task of newsgathering from social media, which has become commonplace for journalists. Given the difficulty of newsgathering from a large data source like social media, there has been a body of research in the use of data mining and natural language processing for facilitating the task, which we have organised in five subsections: news discovery, curation of news, validation and verification of content, newsgathering dashboards, and other tasks. Despite substantial progress in the field, it has proven a difficult task which is still being investigated. The main shortcomings we have identified in present work, which we outline here as important avenues for future research, include:

\begin{itemize}
  \item \textit{\textbf{Development of real-time newsgathering approaches:}} much of the work on newsgathering from social media is conducted post-hoc, i.e. assuming that the entire event has been observed. However, an online or real-time scenario is crucial when newsgathering is being performed for journalists and/or newsrooms. The earliness with which newsworthy reports are identified needs to be an important factor to be measured in future work, making sure that presented approaches are efficiently and effectively applicable in a real-time scenario during breaking news.
  \item \textit{\textbf{Collection and release of datasets:}} while many of the works discussed in this review have collected and annotated their own datasets, few of these are publicly available. Release of datasets is in this case very important to make sure that future research is comparable. Provided the dearth of publicly available datasets, there is a need to generate benchmark datasets for comparison purposes. This can be achieved either by releasing datasets associated with papers, or by running shared tasks where datasets are published for researchers to participate in an open competition. Where a researcher has limited access to a servers to release the data, there are publicly and freely available services such as the public catalogue created as part of the SoBigData project\footnote{\url{https://sobigdata.d4science.org/catalogue-sobigdata}} or data sharing platforms such as Figshare.\footnote{\url{https://figshare.com/}}
  \item \textit{\textbf{Generalisation to new events:}} ideally one wants to test a newsgathering approach in multiple datasets associated with different events, however this is again hindered by the scarcity of publicly available datasets. Release of more datasets should also help generalisation of approach to new events, by training on some events, and testing on other, held-out events. Some of the works above may have a risk of overfitting to a specific dataset, which would improve immensely by experimenting on more datasets. Previous work has looked at generalisation to multiple events in the context of natural disasters \cite{nguyen2017robust,cresci2015linguistically}, which can be used by leveraging techniques for transfer learning or domain adaptation \cite{pan2009survey,ruder2019neural}, however their applications to the context of journalism for newsgathering remains unstudied.
  \item \textit{\textbf{Detection of local news:}} many newsgathering approach rely on outstanding frequencies of vocabulary terms, social activity, etc. While this has proven to be very useful for newsgathering, it also has the limitation that it detects major news events, lacking the ability to detect less popular news that are in the long tail. This is the example of local news, which can be very important in a particular region, but they do not generally sufficient volume on Twitter in order to be detected by the above means. Future research should look into identification of these lesser popular news.
  \item \textit{\textbf{Detection of users in the long tail:}} a similar issue occurs when using approaches for detecting high quality information sources. Depending on the approach, users with many followers, users who tweet more often, etc. may be detected more often. There are, however, witnesses and other high quality users, who do not attract as many retweets, followers,... who are providing useful, unique information. More research is needed in avoiding algorithmic biases that favour popular users in order to be able to find users in the long tail.
  \item \textit{\textbf{Research into fully-fledged newsgathering systems:}} most works above have focused on specific subtasks of the entire pipeline involved in the newsgathering process. Here we have discussed work on event detection, summarisation, news recommenders, content verification, finding information sources and development of dashboards, among others. Integration of these different subtasks into a single, fully-fledged pipeline has not yet been studied.
\end{itemize}

Substantial progress has been made on developing computational approaches to newsgathering from social media by leveraging data mining and natural language processing techniques. While these works have shown encouraging results towards semi-automation of the newsgathering process to assist journalists and news practitioners, there is still more research to be done to deal with the numerous challenges posed by social media. This review has outlined existing research and has identified a set of gaps which are set forth as avenues for future research.

\section*{References}

\bibliographystyle{plain}
\bibliography{smjournalism}

\begin{thebibliography}{100}
\expandafter\ifx\csname url\endcsname\relax
  \def\url#1{\texttt{#1}}\fi
\expandafter\ifx\csname urlprefix\endcsname\relax\def\urlprefix{URL }\fi
\expandafter\ifx\csname href\endcsname\relax
  \def\href#1#2{#2} \def\path#1{#1}\fi

\bibitem{fink2014data}
K.~Fink, Data-Driven Sourcing: How Journalists Use Digital Search Tools to
  Decide What's News, Columbia University, 2014.

\bibitem{fuchs2017social}
C.~Fuchs, Social media: A critical introduction, Sage, 2017.

\bibitem{cohen2011computational}
S.~Cohen, J.~T. Hamilton, F.~Turner, Computational journalism, Communications
  of the ACM 54~(10) (2011) 66--71.

\bibitem{anderson2013towards}
C.~W. Anderson, Towards a sociology of computational and algorithmic
  journalism, new media \& society 15~(7) (2013) 1005--1021.

\bibitem{knight2012journalism}
M.~Knight, Journalism as usual: The use of social media as a newsgathering tool
  in the coverage of the iranian elections in 2009, Journal of Media Practice
  13~(1) (2012) 61--74.

\bibitem{knight2013social}
M.~Knight, C.~Cook, Social media for journalists: Principles and practice, SAGE
  Publications Limited, 2013.

\bibitem{belair2015social}
V.~Belair-Gagnon, Social media at BBC news: The re-making of crisis reporting,
  Vol.~10, Routledge, 2015.

\bibitem{abdenour2017digital}
J.~Abdenour, Digital gumshoes: Investigative journalists’ use of social media
  in television news reporting, Digital Journalism 5~(4) (2017) 472--492.

\bibitem{lysak2012facebook}
S.~Lysak, M.~Cremedas, J.~Wolf, Facebook and twitter in the newsroom: How and
  why local television news is getting social with viewers?, Electronic News
  6~(4) (2012) 187--207.

\bibitem{murrell2014vulture}
C.~Murrell, et~al., The vulture club: International newsgathering via facebook,
  Australian Journalism Review 36~(1) (2014) 15.

\bibitem{kaonza2015social}
C.~V. Kaonza, Social media, a handtool for news gathering?: a case study of the
  newsday and southern eye.

\bibitem{zakaria2018effect}
N.~A. Zakaria, F.~H.~A. Razak, The effect of facebook on journalist in news
  writing, Journal of Media and Information Warfare Vol 11 (2018) 29--49.

\bibitem{ighiegba2018utilisation}
O.~S. Ighiegba, W.~O. OLLEY, Utilisation of social media for news gathering and
  dissemination by journalists in edo state, nigeria, The Nigerian Journal of
  Communication (TNJC) 15~(2).

\bibitem{fichet2016eyes}
E.~S. Fichet, J.~J. Robinson, D.~Dailey, K.~Starbird, Eyes on the ground:
  Emerging practices in periscope use during crisis events., in: ISCRAM, 2016,
  pp. 1--10.

\bibitem{tang2016meerkat}
J.~C. Tang, G.~Venolia, K.~M. Inkpen, Meerkat and periscope: I stream, you
  stream, apps stream for live streams, in: Proceedings of the 2016 CHI
  conference on human factors in computing systems, ACM, 2016, pp. 4770--4780.

\bibitem{alejandro2010journalism}
J.~Alejandro, Journalism in the age of social media, Reuters Institute
  Fellowship Paper, University of Oxford (2010) 2009--2010.

\bibitem{hermida2012social}
A.~Hermida, Social journalism: Exploring how social media is shaping
  journalism, The handbook of global online journalism (2012) 309--328.

\bibitem{broersma2013twitter}
M.~Broersma, T.~Graham, Twitter as a news source: How dutch and british
  newspapers used tweets in their news coverage, 2007--2011, Journalism
  Practice 7~(4) (2013) 446--464.

\bibitem{hermida2013journalism}
A.~Hermida, \# journalism: Reconfiguring journalism research about twitter, one
  tweet at a time, Digital Journalism 1~(3) (2013) 295--313.

\bibitem{phuvipadawat2010breaking}
S.~Phuvipadawat, T.~Murata, Breaking news detection and tracking in twitter,
  in: Proceedings of the 2010 IEEE/WIC/ACM International Conference on Web
  Intelligence and Intelligent Agent Technology (WI-IAT), IEEE, 2010, pp.
  120--123.

\bibitem{hu2012breaking}
M.~Hu, S.~Liu, F.~Wei, Y.~Wu, J.~Stasko, K.-L. Ma, Breaking news on twitter,
  in: Proceedings of the SIGCHI Conference on Human Factors in Computing
  Systems, ACM, 2012, pp. 2751--2754.

\bibitem{gross2017harvesting}
B.~Gross, Harvesting social media for journalistic purposes in the uk, in:
  Privacy, Data Protection and Cybersecurity in Europe, Springer, 2017, pp.
  31--42.

\bibitem{brands2018social}
B.~J. Brands, T.~Graham, M.~Broersma, Social media sourcing practices: How
  dutch newspapers use tweets in political news coverage, in: Managing
  Democracy in the Digital Age, Springer, 2018, pp. 159--178.

\bibitem{lewis2019social}
S.~Lewis, L.~Molyneux, Social media and journalism: 10 years later, untangling
  key assumptions, in: Proceedings of the 52nd Hawaii International Conference
  on System Sciences, 2019.

\bibitem{priya2019should}
S.~Priya, R.~Sequeira, J.~Chandra, S.~K. Dandapat, Where should one get news
  updates: Twitter or reddit, Online Social Networks and Media 9 (2019) 17--29.

\bibitem{lenhart2009twitter}
A.~Lenhart, S.~Fox, Twitter and status updating, Pew Internet \& American Life
  Project Washington, DC, 2009.

\bibitem{khondker2011role}
H.~H. Khondker, Role of the new media in the arab spring, Globalizations 8~(5)
  (2011) 675--679.

\bibitem{orellana2018attention}
C.~Orellana-Rodriguez, M.~T. Keane, Attention to news and its dissemination on
  twitter: A survey, Computer Science Review 29 (2018) 74--94.

\bibitem{zubiaga2019social}
A.~Zubiaga, B.~Heravi, J.~An, H.~Kwak, Social media mining for journalism,
  Online Information Review 43~(1) (2019) 2--6.

\bibitem{saldana2017sharing}
M.~Salda{\~n}a, V.~d.~M. Higgins~Joyce, A.~Schmitz~Weiss, R.~C. Alves, Sharing
  the stage: Analysis of social media adoption by latin american journalists,
  Journalism Practice 11~(4) (2017) 396--416.

\bibitem{shoenberger2017perceived}
H.~Shoenberger, S.~Rodgers, Perceived health reporting knowledge and news
  gathering practices of health journalists and editors at community
  newspapers, Journal of health communication 22~(3) (2017) 205--213.

\bibitem{li2017better}
B.~Li, S.~Stokowski, S.~W. Dittmore, O.~K. Scott, For better or for worse: The
  impact of social media on chinese sports journalists, Communication \& Sport
  5~(3) (2017) 311--330.

\bibitem{bowdon2014tweeting}
M.~A. Bowdon, Tweeting an ethos: Emergency messaging, social media, and
  teaching technical communication, Technical Communication Quarterly 23~(1)
  (2014) 35--54.

\bibitem{parmelee2013political}
J.~H. Parmelee, Political journalists and twitter: Influences on norms and
  practices, Journal of Media Practice 14~(4) (2013) 291--305.

\bibitem{kwak2010twitter}
H.~Kwak, C.~Lee, H.~Park, S.~Moon, What is twitter, a social network or a news
  media?, in: Proceedings of the 19th international conference on World wide
  web, ACM, 2010, pp. 591--600.

\bibitem{diakopoulos2014newsworthiness}
N.~Diakopoulos, A.~Zubiaga, Newsworthiness and network gatekeeping on twitter:
  The role of social deviance, in: Proceedings of the International Conference
  on Weblogs and Social Media, 2014, pp. 587--590.

\bibitem{diakopoulos2012finding}
N.~Diakopoulos, M.~De~Choudhury, M.~Naaman, Finding and assessing social media
  information sources in the context of journalism, in: Proceedings of the
  SIGCHI Conference on Human Factors in Computing Systems, ACM, 2012, pp.
  2451--2460.

\bibitem{imran2013extracting}
M.~Imran, S.~Elbassuoni, C.~Castillo, F.~Diaz, P.~Meier, Extracting information
  nuggets from disaster-related messages in social media, in: Proceedings of
  ISCRAM, 2013, pp. 791--800.

\bibitem{zubiaga2013curating}
A.~Zubiaga, H.~Ji, K.~Knight, Curating and contextualizing twitter stories to
  assist with social newsgathering, in: Proceedings of the 2013 international
  conference on Intelligent user interfaces, ACM, 2013, pp. 213--224.

\bibitem{tolmie2017supporting}
P.~Tolmie, R.~Procter, D.~W. Randall, M.~Rouncefield, C.~Burger, G.~Wong
  Sak~Hoi, A.~Zubiaga, M.~Liakata, Supporting the use of user generated content
  in journalistic practice, in: Proceedings of the 2017 CHI Conference on Human
  Factors in Computing Systems, ACM, 2017, pp. 3632--3644.

\bibitem{thurman2018social}
N.~Thurman, Social media, surveillance, and news work: On the apps promising
  journalists a “crystal ball”, Digital Journalism 6~(1) (2018) 76--97.

\bibitem{hermida2010tv}
A.~Hermida, From tv to twitter: How ambient news became ambient journalism,
  Media/Culture Journal 13~(2).

\bibitem{dou2012event}
W.~Dou, K.~Wang, W.~Ribarsky, M.~Zhou, Event detection in social media data,
  in: IEEE VisWeek Workshop on Interactive Visual Text Analytics-Task Driven
  Analytics of Social Media Content, 2012, pp. 971--980.

\bibitem{abel2012twitcident}
F.~Abel, C.~Hauff, G.-J. Houben, R.~Stronkman, K.~Tao, Twitcident: fighting
  fire with information from social web streams, in: Proceedings of the 21st
  International Conference on World Wide Web, ACM, 2012, pp. 305--308.

\bibitem{schulz2013see}
A.~Schulz, P.~Ristoski, H.~Paulheim, I see a car crash: Real-time detection of
  small scale incidents in microblogs, in: Proceedings of the Extended Semantic
  Web Conference, Springer, 2013, pp. 22--33.

\bibitem{weiler2013event}
A.~Weiler, M.~H. Scholl, F.~Wanner, C.~Rohrdantz, Event identification for
  local areas using social media streaming data, in: Proceedings of the ACM
  SIGMOD Workshop on Databases and Social Networks, ACM, 2013, pp. 1--6.

\bibitem{wei2017finding}
H.~Wei, J.~Sankaranarayanan, H.~Samet, Finding and tracking local twitter users
  for news detection, in: Proceedings of the 25th ACM SIGSPATIAL International
  Conference on Advances in Geographic Information Systems, 2017.

\bibitem{xie2016topicsketch}
W.~Xie, F.~Zhu, J.~Jiang, E.-P. Lim, K.~Wang, Topicsketch: Real-time bursty
  topic detection from twitter, IEEE Transactions on Knowledge and Data
  Engineering 28~(8) (2016) 2216--2229.

\bibitem{liu2016reuters}
X.~Liu, Q.~Li, A.~Nourbakhsh, R.~Fang, M.~Thomas, K.~Anderson, R.~Kociuba,
  M.~Vedder, S.~Pomerville, R.~Wudali, et~al., Reuters tracer: A large scale
  system of detecting \& verifying real-time news events from twitter, in:
  Proceedings of the 25th ACM International on Conference on Information and
  Knowledge Management, ACM, 2016, pp. 207--216.

\bibitem{sakaki2010earthquake}
T.~Sakaki, M.~Okazaki, Y.~Matsuo, Earthquake shakes twitter users: real-time
  event detection by social sensors, in: Proceedings of the 19th international
  conference on World wide web, ACM, 2010, pp. 851--860.

\bibitem{rogstadius2013crisistracker}
J.~Rogstadius, M.~Vukovic, C.~A. Teixeira, V.~Kostakos, E.~Karapanos, J.~A.
  Laredo, Crisistracker: Crowdsourced social media curation for disaster
  awareness, IBM Journal of Research and Development 57~(5) (2013) 4--1.

\bibitem{middleton2014real}
S.~E. Middleton, L.~Middleton, S.~Modafferi, Real-time crisis mapping of
  natural disasters using social media, IEEE Intelligent Systems 29~(2) (2014)
  9--17.

\bibitem{avvenuti2014ears}
M.~Avvenuti, S.~Cresci, A.~Marchetti, C.~Meletti, M.~Tesconi, Ears (earthquake
  alert and report system): a real time decision support system for earthquake
  crisis management, in: Proceedings of the 20th ACM SIGKDD international
  conference on knowledge discovery and data mining, ACM, 2014, pp. 1749--1758.

\bibitem{yin2015using}
J.~Yin, S.~Karimi, A.~Lampert, M.~Cameron, B.~Robinson, R.~Power, Using social
  media to enhance emergency situation awareness, in: Twenty-Fourth
  International Joint Conference on Artificial Intelligence, 2015.

\bibitem{maresh2016social}
M.~M. Maresh-Fuehrer, R.~Smith, Social media mapping innovations for crisis
  prevention, response, and evaluation, Computers in Human Behavior 54 (2016)
  620--629.

\bibitem{avvenuti2016impromptu}
M.~Avvenuti, S.~Cresci, F.~Del~Vigna, M.~Tesconi, Impromptu crisis mapping to
  prioritize emergency response, Computer 49~(5) (2016) 28--37.

\bibitem{imran2018processing}
M.~Imran, C.~Castillo, F.~Diaz, S.~Vieweg, Processing social media messages in
  mass emergency: Survey summary, in: Companion Proceedings of the The Web
  Conference 2018, International World Wide Web Conferences Steering Committee,
  2018, pp. 507--511.

\bibitem{toujani2019event}
R.~Toujani, J.~Akaichi, Event news detection and citizens community structure
  for disaster management in social networks, Online Information Review 43~(1)
  (2019) 113--132.

\bibitem{asur2011trends}
S.~Asur, B.~A. Huberman, G.~Szabo, C.~Wang, Trends in social media: Persistence
  and decay, in: Fifth international AAAI conference on weblogs and social
  media, 2011.

\bibitem{becker2011beyond}
H.~Becker, M.~Naaman, L.~Gravano, Beyond trending topics: Real-world event
  identification on twitter, in: Fifth international AAAI conference on weblogs
  and social media, 2011.

\bibitem{aiello2013sensing}
L.~M. Aiello, G.~Petkos, C.~Martin, D.~Corney, S.~Papadopoulos, R.~Skraba,
  A.~G{\"o}ker, I.~Kompatsiaris, A.~Jaimes, Sensing trending topics in twitter,
  IEEE Transactions on Multimedia 15~(6) (2013) 1268--1282.

\bibitem{pervin2013fast}
N.~Pervin, F.~Fang, A.~Datta, K.~Dutta, D.~Vandermeer, Fast, scalable, and
  context-sensitive detection of trending topics in microblog post streams, ACM
  transactions on management information systems (TMIS) 3~(4) (2013) 19.

\bibitem{zubiaga2015real}
A.~Zubiaga, D.~Spina, R.~Mart{\'\i}nez, V.~Fresno, Real-time classification of
  twitter trends, Journal of the Association for Information Science and
  Technology 66~(3) (2015) 462--473.

\bibitem{naveed2019location}
N.~Naveed, M.~Abbas, Z.~Rauf, Location based sentiment mapping of topics
  detected in social media, Journal of Applied and Emerging Sciences 8~(2)
  (2019) pp108--117.

\bibitem{atefeh2015survey}
F.~Atefeh, W.~Khreich, A survey of techniques for event detection in twitter,
  Computational Intelligence 31~(1) (2015) 132--164.

\bibitem{phelan2009using}
O.~Phelan, K.~McCarthy, B.~Smyth, Using twitter to recommend real-time topical
  news, in: Proceedings of the third ACM conference on Recommender systems,
  ACM, 2009, pp. 385--388.

\bibitem{phelan2011terms}
O.~Phelan, K.~McCarthy, M.~Bennett, B.~Smyth, Terms of a feather: Content-based
  news recommendation and discovery using twitter, Advances in Information
  Retrieval (2011) 448--459.

\bibitem{krestel2015tweet}
R.~Krestel, T.~Werkmeister, T.~P. Wiradarma, G.~Kasneci, Tweet-recommender:
  Finding relevant tweets for news articles, in: Proceedings of the 24th
  International Conference on World Wide Web, ACM, 2015, pp. 53--54.

\bibitem{cucchiarelli2017write}
A.~Cucchiarelli, C.~Morbidoni, G.~Stilo, P.~Velardi, What to write? a topic
  recommender for journalists, in: Proceedings of the 2017 EMNLP Workshop:
  Natural Language Processing meets Journalism, 2017, pp. 19--24.

\bibitem{cucchiarelli2019topic}
A.~Cucchiarelli, C.~Morbidoni, G.~Stilo, P.~Velardi, A topic recommender for
  journalists, Information Retrieval Journal 22~(1-2) (2019) 4--31.

\bibitem{stilo2017capturing}
G.~Stilo, C.~Morbidoni, A.~Cucchiarelli, P.~Velardi, Capturing users’
  information and communication needs for the press officers (2017) 14--27.

\bibitem{zubiaga2012towards}
A.~Zubiaga, D.~Spina, E.~Amig{\'o}, J.~Gonzalo, Towards real-time summarization
  of scheduled events from twitter streams, in: Proceedings of the 23rd ACM
  conference on Hypertext and social media, ACM, 2012, pp. 319--320.

\bibitem{nichols2012summarizing}
J.~Nichols, J.~Mahmud, C.~Drews, Summarizing sporting events using twitter, in:
  Proceedings of the 2012 ACM international conference on Intelligent User
  Interfaces, ACM, 2012, pp. 189--198.

\bibitem{khan2013multi}
M.~A.~H. Khan, D.~Bollegala, G.~Liu, K.~Sezaki, Multi-tweet summarization of
  real-time events, in: Social Computing (SocialCom), 2013 International
  Conference on, IEEE, 2013, pp. 128--133.

\bibitem{meladianos2018optimization}
P.~Meladianos, C.~Xypolopoulos, G.~Nikolentzos, M.~Vazirgiannis, An
  optimization approach for sub-event detection and summarization in twitter,
  in: European Conference on Information Retrieval, Springer, 2018, pp.
  481--493.

\bibitem{wang2015summarization}
Z.~Wang, L.~Shou, K.~Chen, G.~Chen, S.~Mehrotra, On summarization and timeline
  generation for evolutionary tweet streams, IEEE Transactions on Knowledge and
  Data Engineering 27~(5) (2015) 1301--1315.

\bibitem{shapira2017interactive}
O.~Shapira, H.~Ronen, M.~Adler, Y.~Amsterdamer, J.~Bar-Ilan, I.~Dagan,
  Interactive abstractive summarization for event news tweets, in: Proceedings
  of the 2017 Conference on Empirical Methods in Natural Language Processing:
  System Demonstrations, 2017, pp. 109--114.

\bibitem{starbird2012learning}
K.~Starbird, G.~Muzny, L.~Palen, Learning from the crowd: collaborative
  filtering techniques for identifying on-the-ground twitterers during mass
  disruptions, in: Proceedings of 9th International Conference on Information
  Systems for Crisis Response and Management, 2012, pp. 1--10.

\bibitem{morstatter2014finding}
F.~Morstatter, N.~Lubold, H.~Pon-Barry, J.~Pfeffer, H.~Liu, Finding eyewitness
  tweets during crises, in: Proceedings of the ACL 2014 Workshop on Language
  Technologies and Computational Social Science, 2014, pp. 23--27.

\bibitem{mahata2012does}
D.~Mahata, N.~Agarwal, What does everybody know? identifying event-specific
  sources from social media, in: Computational Aspects of Social Networks
  (CASoN), 2012 Fourth International Conference on, IEEE, 2012, pp. 63--68.

\bibitem{fang2016witness}
R.~Fang, A.~Nourbakhsh, X.~Liu, S.~Shah, Q.~Li, Witness identification in
  twitter, in: Proceedings of the Fourth International Workshop on Natural
  Language Processing for Social Media, 2016, pp. 65--73.

\bibitem{doggett2016identifying}
E.~V. Doggett, A.~Cantarero, Identifying eyewitness news-worthy events on
  twitter, in: Proceedings of the Conference on Empirical Methods in Natural
  Language Processing, 2016, pp. 7--13.

\bibitem{truelove2014testing}
M.~Truelove, M.~Vasardani, S.~Winter, Testing a model of witness accounts in
  social media, in: Proceedings of the 8th Workshop on Geographic Information
  Retrieval, ACM, 2014, pp. 10:1--10:8.

\bibitem{truelove2017identifying}
M.~Truelove, K.~Khoshelham, S.~McLean, S.~Winter, M.~Vasardani, Identifying
  witness accounts from social media using imagery, ISPRS International Journal
  of Geo-Information 6~(4) (2017) 120:1--120:24.

\bibitem{truelove2017testing}
M.~Truelove, M.~Vasardani, S.~Winter, Testing the event witnessing status of
  micro-bloggers from evidence in their micro-blogs, PloS one 12~(12) (2017)
  e0189378.

\bibitem{silverman2014verification}
C.~Silverman, Verification handbook: a definitive guide to verifying digital
  content for emergency coverage, European Journalism Centre, 2014.

\bibitem{gupta2013faking}
A.~Gupta, H.~Lamba, P.~Kumaraguru, A.~Joshi, Faking sandy: characterizing and
  identifying fake images on twitter during hurricane sandy, in: Proceedings of
  the 22nd international conference on World Wide Web, ACM, 2013, pp. 729--736.

\bibitem{schifferes2014identifying}
S.~Schifferes, N.~Newman, N.~Thurman, D.~Corney, A.~G{\"o}ker, C.~Martin,
  Identifying and verifying news through social media: Developing a
  user-centred tool for professional journalists, Digital Journalism 2~(3)
  (2014) 406--418.

\bibitem{brandtzaeg2016emerging}
P.~B. Brandtzaeg, M.~L{\"u}ders, J.~Spangenberg, L.~Rath-Wiggins,
  A.~F{\o}lstad, Emerging journalistic verification practices concerning social
  media, Journalism Practice 10~(3) (2016) 323--342.

\bibitem{fletcher2017building}
R.~Fletcher, S.~Schifferes, N.~Thurman, Building the ‘truthmeter’ training
  algorithms to help journalists assess the credibility of social media
  sources, Convergence (2017) 1354856517714955.

\bibitem{lewandowsky2012misinformation}
S.~Lewandowsky, U.~K. Ecker, C.~M. Seifert, N.~Schwarz, J.~Cook, Misinformation
  and its correction: Continued influence and successful debiasing,
  Psychological Science in the Public Interest 13~(3) (2012) 106--131.

\bibitem{zubiaga2014tweet}
A.~Zubiaga, H.~Ji, Tweet, but verify: epistemic study of information
  verification on twitter, Social Network Analysis and Mining 4~(1) (2014)
  1--12.

\bibitem{liu2015real}
X.~Liu, A.~Nourbakhsh, Q.~Li, R.~Fang, S.~Shah, Real-time rumor debunking on
  twitter, in: Proceedings of the 24th ACM International on Conference on
  Information and Knowledge Management, ACM, 2015, pp. 1867--1870.

\bibitem{yang2012automatic}
F.~Yang, Y.~Liu, X.~Yu, M.~Yang, Automatic detection of rumor on sina weibo,
  in: Proceedings of the ACM SIGKDD Workshop on Mining Data Semantics, ACM,
  2012, p.~13.

\bibitem{castillo2011information}
C.~Castillo, M.~Mendoza, B.~Poblete, Information credibility on twitter, in:
  Proceedings of the 20th international conference on World wide web, ACM,
  2011, pp. 675--684.

\bibitem{vosoughi2015automatic}
S.~Vosoughi, Automatic detection and verification of rumors on twitter, Ph.D.
  thesis (2015).

\bibitem{boididou2017verifying}
C.~Boididou, S.~E. Middleton, Z.~Jin, S.~Papadopoulos, D.-T. Dang-Nguyen,
  G.~Boato, Y.~Kompatsiaris, Verifying information with multimedia content on
  twitter, Multimedia Tools and Applications (2017) 1--27.

\bibitem{derczynski2017semeval}
L.~Derczynski, K.~Bontcheva, M.~Liakata, R.~Procter, G.~Wong Sak~Hoi,
  A.~Zubiaga, {SemEval-2017 Task 8: RumourEval: Determining rumour veracity and
  support for rumours}, in: Proceedings of SemEval, ACL, 2017.

\bibitem{enayet2017niletmrg}
O.~Enayet, S.~R. El-Beltagy, {NileTMRG at SemEval-2017 Task 8: Determining
  Rumour and Veracity Support for Rumours on Twitter}, in: Proceedings of
  SemEval, ACL, 2017.

\bibitem{wu2014toward}
Y.~Wu, P.~K. Agarwal, C.~Li, J.~Yang, C.~Yu, Toward computational
  fact-checking, Proceedings of the VLDB Endowment 7~(7) (2014) 589--600.

\bibitem{ciampaglia2015computational}
G.~L. Ciampaglia, P.~Shiralkar, L.~M. Rocha, J.~Bollen, F.~Menczer,
  A.~Flammini, Computational fact checking from knowledge networks, PloS one
  10~(6) (2015) e0128193.

\bibitem{shiralkar2017computational}
P.~Shiralkar, Computational fact checking by mining knowledge graphs, Ph.D.
  thesis, Indiana University (2017).

\bibitem{shi2016discriminative}
B.~Shi, T.~Weninger, Discriminative predicate path mining for fact checking in
  knowledge graphs, Knowledge-Based Systems 104 (2016) 123--133.

\bibitem{shao2016hoaxy}
C.~Shao, G.~L. Ciampaglia, A.~Flammini, F.~Menczer, Hoaxy: A platform for
  tracking online misinformation, in: Proceedings of the 25th International
  Conference Companion on World Wide Web, International World Wide Web
  Conferences Steering Committee, 2016, pp. 745--750.

\bibitem{boididou2015verifying}
C.~Boididou, K.~Andreadou, S.~Papadopoulos, D.-T. Dang-Nguyen, G.~Boato,
  M.~Riegler, Y.~Kompatsiaris, et~al., Verifying multimedia use at mediaeval
  2015., in: Proceedings of MediaEval, 2015.

\bibitem{zampoglou2016web}
M.~Zampoglou, S.~Papadopoulos, Y.~Kompatsiaris, R.~Bouwmeester, J.~Spangenberg,
  Web and social media image forensics for news professionals, in: Proceedings
  of the Workshop on Social Media in the Newsroom, 2016, pp. 159--166.

\bibitem{middleton2016geoparsing}
S.~E. Middleton, V.~Krivcovs, Geoparsing and geosemantics for social media:
  spatiotemporal grounding of content propagating rumors to support trust and
  veracity analysis during breaking news, ACM Transactions on Information
  Systems (TOIS) 34~(3) (2016) 16.

\bibitem{middleton2018social}
S.~Middleton, S.~Papadopoulos, Y.~Kompatsiaris, Social computing for verifying
  social media content in breaking news, IEEE Internet Computing.

\bibitem{andreadou2015media}
K.~Andreadou, S.~Papadopoulos, L.~Apostolidis, A.~Krithara, Y.~Kompatsiaris,
  Media revealr: A social multimedia monitoring and intelligence system for web
  multimedia verification, in: Pacific-Asia Workshop on Intelligence and
  Security Informatics, Springer, 2015, pp. 1--20.

\bibitem{papadopoulou2017web}
O.~Papadopoulou, M.~Zampoglou, S.~Papadopoulos, Y.~Kompatsiaris, Web video
  verification using contextual cues, in: Proceedings of the 2nd International
  Workshop on Multimedia Forensics and Security, ACM, 2017, pp. 6--10.

\bibitem{zubiaga2018detection}
A.~Zubiaga, A.~Aker, K.~Bontcheva, M.~Liakata, R.~Procter, Detection and
  resolution of rumours in social media: A survey, ACM Computing Surveys (CSUR)
  51~(2) (2018) 32:1--32:36.

\bibitem{shu2017fake}
K.~Shu, A.~Sliva, S.~Wang, J.~Tang, H.~Liu, Fake news detection on social
  media: A data mining perspective, ACM SIGKDD Explorations Newsletter 19~(1)
  (2017) 22--36.

\bibitem{marcus2011twitinfo}
A.~Marcus, M.~S. Bernstein, O.~Badar, D.~R. Karger, S.~Madden, R.~C. Miller,
  Twitinfo: aggregating and visualizing microblogs for event exploration, in:
  Proceedings of the SIGCHI conference on Human factors in computing systems,
  ACM, 2011, pp. 227--236.

\bibitem{lin2016information}
C.-Y. Lin, T.-Y. Li, P.~Chen, An information visualization system to assist
  news topics exploration with social media, in: Proceedings of the 7th 2016
  International Conference on Social Media \& Society, ACM, 2016, pp.
  23:1--23:9.

\bibitem{amer2012maqsa}
S.~Amer-Yahia, S.~Anjum, A.~Ghenai, A.~Siddique, S.~Abbar, S.~Madden,
  A.~Marcus, M.~El-Haddad, Maqsa: a system for social analytics on news, in:
  Proceedings of the 2012 ACM SIGMOD International Conference on Management of
  Data, ACM, 2012, pp. 653--656.

\bibitem{schwartz2015editorial}
R.~Schwartz, M.~Naaman, R.~Teodoro, Editorial algorithms: Using social media to
  discover and report local news., in: Proceedings of the International
  Conference on Web and Social Media, 2015, pp. 407--415.

\bibitem{milajevs2013real}
D.~Milajevs, G.~Bouma, Real time discussion retrieval from twitter, in:
  Proceedings of the 22nd International Conference on World Wide Web, ACM,
  2013, pp. 795--800.

\bibitem{wang2013exploiting}
X.~Wang, L.~Tokarchuk, F.~Cuadrado, S.~Poslad, Exploiting hashtags for adaptive
  microblog crawling, in: Proceedings of the 2013 IEEE/ACM international
  conference on advances in social networks analysis and mining, ACM, 2013, pp.
  311--315.

\bibitem{wang2014identifying}
X.~Wang, L.~Tokarchuk, S.~Poslad, Identifying relevant event content for
  real-time event detection, in: Advances in Social Networks Analysis and
  Mining (ASONAM), 2014 IEEE/ACM International Conference on, IEEE, 2014, pp.
  395--398.

\bibitem{wang2015adaptive}
X.~Wang, L.~Tokarchuk, F.~Cuadrado, S.~Poslad, Adaptive identification of
  hashtags for real-time event data collection, in: Recommendation and Search
  in Social Networks, Springer, 2015, pp. 1--22.

\bibitem{fresno2015exploiting}
V.~Fresno, A.~Zubiaga, H.~Ji, R.~Mart{\'\i}nez, Exploiting geolocation, user
  and temporal information for natural hazards monitoring in twitter,
  Procesamiento del Lenguaje Natural 54 (2015) 85--92.

\bibitem{freitas2016identifying}
J.~Freitas, H.~Ji, Identifying news from tweets, in: Proceedings of 2016 EMNLP
  Workshop on Natural Language Processing and Computational Social Science,
  2016, pp. 11--16.

\bibitem{stowe2016identifying}
K.~Stowe, M.~Paul, M.~Palmer, L.~Palen, K.~Anderson, Identifying and
  categorizing disaster-related tweets, in: Proceedings of The Fourth
  International Workshop on Natural Language Processing for Social Media, 2016,
  pp. 1--6.

\bibitem{lukasik2016hawkes}
M.~Lukasik, P.~Srijith, D.~Vu, K.~Bontcheva, A.~Zubiaga, T.~Cohn, Hawkes
  processes for continuous time sequence classification: an application to
  rumour stance classification in twitter, in: Proceedings of the 54th Annual
  Meeting of the Association for Computational Linguistics (Volume 2: Short
  Papers), Vol.~2, 2016, pp. 393--398.

\bibitem{zubiaga2018discourse}
A.~Zubiaga, E.~Kochkina, M.~Liakata, R.~Procter, M.~Lukasik, K.~Bontcheva,
  T.~Cohn, I.~Augenstein, Discourse-aware rumour stance classification in
  social media using sequential classifiers, Information Processing \&
  Management 54~(2) (2018) 273--290.

\bibitem{petrovic2010streaming}
S.~Petrovi{\'c}, M.~Osborne, V.~Lavrenko, Streaming first story detection with
  application to twitter, in: Human language technologies: The 2010 annual
  conference of the north american chapter of the association for computational
  linguistics, Association for Computational Linguistics, 2010, pp. 181--189.

\bibitem{petrovic2012using}
S.~Petrovi{\'c}, M.~Osborne, V.~Lavrenko, Using paraphrases for improving first
  story detection in news and twitter, in: Proceedings of the 2012 conference
  of the north american chapter of the association for computational
  linguistics: Human language technologies, Association for Computational
  Linguistics, 2012, pp. 338--346.

\bibitem{osborne2012bieber}
M.~Osborne, S.~Petrovic, R.~McCreadie, C.~Macdonald, I.~Ounis, Bieber no more:
  First story detection using twitter and wikipedia, in: Sigir 2012 workshop on
  time-aware information access, 2012.

\bibitem{babaei2018purple}
M.~Babaei, J.~Kulshrestha, A.~Chakraborty, F.~Benevenuto, K.~P. Gummadi,
  A.~Weller, Purple feed: Identifying high consensus news posts on social
  media, in: Proceedings of the AAAI/ACM conference on Artificial Intelligence,
  Ethics and Society, 2018.

\bibitem{nguyen2017robust}
D.~T. Nguyen, K.~A. Al~Mannai, S.~Joty, H.~Sajjad, M.~Imran, P.~Mitra, Robust
  classification of crisis-related data on social networks using convolutional
  neural networks, in: Eleventh International AAAI Conference on Web and Social
  Media, 2017.

\bibitem{cresci2015linguistically}
S.~Cresci, M.~Tesconi, A.~Cimino, F.~Dell'Orletta, A linguistically-driven
  approach to cross-event damage assessment of natural disasters from social
  media messages, in: Proceedings of the 24th International Conference on World
  Wide Web, ACM, 2015, pp. 1195--1200.

\bibitem{pan2009survey}
S.~J. Pan, Q.~Yang, A survey on transfer learning, IEEE Transactions on
  knowledge and data engineering 22~(10) (2009) 1345--1359.

\bibitem{ruder2019neural}
S.~Ruder, Neural transfer learning for natural language processing, Ph.D.
  thesis, National University of Ireland, Galway (2019).

\end{thebibliography}

\end{document}